\documentclass{article}
\usepackage[table,xcdraw,dvipsnames]{xcolor}
\definecolor{darkgreen}{rgb}{0.0, 0.5, 0.0} 

\usepackage{microtype}
\usepackage{graphicx}
\usepackage{subfigure}
\usepackage{booktabs} 
\usepackage{graphicx} 
\usepackage{relsize}  
\usepackage{hyperref}
\usepackage{booktabs}            
\usepackage{graphicx}            
\usepackage{amssymb}             
\usepackage[section]{placeins}

\usepackage[accepted]{icml2025}
\usepackage{amsmath}
\usepackage{amssymb}
\usepackage{mathtools}
\usepackage{amsthm}
\usepackage{booktabs} 
\usepackage{multirow} 
\usepackage[capitalize,noabbrev]{cleveref}
\usepackage{hyperref}
\hypersetup{
hidelinks,
colorlinks=true,
allcolors=black,
pdfstartview=Fit,
breaklinks=true
}
\theoremstyle{plain}

\theoremstyle{definition}

\theoremstyle{remark}

\usepackage[textsize=tiny]{todonotes}

\icmltitlerunning{}

\begin{document}

\twocolumn[
\icmltitle{From Visuals to Vocabulary: Establishing Equivalence Between\\ Image and Text Token Through Autoregressive Pre-training in MLLMs}



\icmlsetsymbol{equal}{*}

\begin{icmlauthorlist}
\icmlauthor{Mingxiao Li}{equal,hunyuan}
\icmlauthor{Fang Qu}{equal,ustc}
\icmlauthor{Zhanpeng Chen}{hunyuan}
\icmlauthor{Na Su}{wxg}
\icmlauthor{Zhizhou Zhong}{fudan}
\icmlauthor{Ziyang Chen}{hunyuan}
\\
\icmlauthor{Nan Du}{hunyuan}
\icmlauthor{Xiaolong Li}{hunyuan}
\end{icmlauthorlist}

\icmlaffiliation{hunyuan}{Tencent Hunyuan}
\icmlaffiliation{wxg}{Tencent WXG Group}
\icmlaffiliation{fudan}{Fudan University}
\icmlaffiliation{ustc}{University of Science and Technology of China}


\icmlkeywords{Machine Learning, ICML}

\vskip 0.3in
]



\printAffiliationsAndNotice{\icmlEqualContribution} 

\begin{abstract}
While MLLMs perform well on perceptual tasks, they lack precise multimodal alignment, limiting performance. 
To address this challenge, we propose Vision Dynamic Embedding-Guided Pretraining (VDEP), a hybrid autoregressive training paradigm for MLLMs. Utilizing dynamic embeddings from the MLP following the visual encoder, this approach supervises image hidden states and integrates image tokens into autoregressive training. 
Existing MLLMs primarily focused on recovering information from textual inputs, often neglecting the effective processing of image data. In contrast,  the key improvement of this work is the reinterpretation of multimodal alignment as a process of recovering information from input data, with particular emphasis on reconstructing detailed visual features.
The proposed method seamlessly integrates into standard models without architectural changes. Experiments on 13 benchmarks show VDEP outperforms baselines, surpassing existing methods.
\end{abstract}
\section{Introduction}
\label{introduction}

The advent of large language models (LLMs) such as ChatGPT \cite{schulman2022chatgpt} has profoundly reshaped the trajectory of AGI development, showcasing exceptional zero-shot reasoning capabilities in addressing various NLP tasks via user-defined prompts or language instructions. 
Traditional vision foundation models typically follow a pre-training and fine-tuning paradigm \cite{wang2023internimage, chen2022vision, su2023towards, wang2023image, tao2023siamese}. While effective, this approach incurs significant marginal costs when adapting to diverse downstream tasks.
In contrast, LLMs open new opportunities for developing unified frameworks to address visual tasks. Multimodal large language models (MLLMs) extend this paradigm by employing vision pre-training models to project image features into a shared textual representation space, leveraging the language instruction capabilities of LLMs to adapt to a wide range of vision tasks. 
These models have achieved state-of-the-art results across multiple application domains, including information retrieval, video analysis, and open-world question answering, as evidenced by their superior benchmark scores and qualitative outputs. 
\begin{figure*}[ht]
\vskip 0.2in
\begin{center}
\centerline{\includegraphics[width=0.9\textwidth]{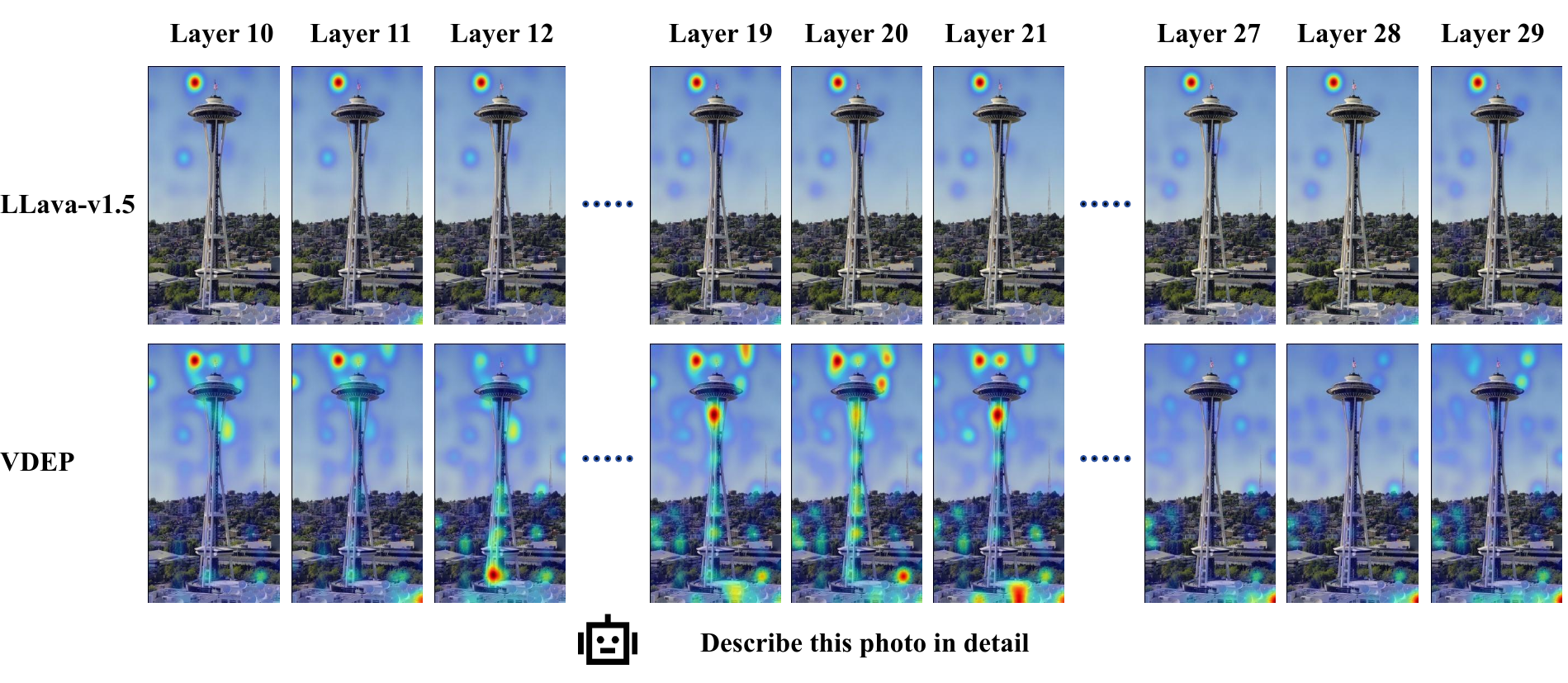}}
\caption{Layer-wise attention visualization of visual-to-instruction information flow. The
example is derived from LLava-Bench \cite{liu2024improved} and the query is "Describe this photo in detail". The visualization results demonstrate that VDEP significantly enhances the model's ability to capture critical features in images, with particularly outstanding performance in identifying object boundaries.}
\label{lighthouse}
\end{center}
\vskip -0.2in
\end{figure*}

However, MLLMs face significant challenges, particularly in multimodal fusion and alignment. Current MLLMs rely heavily on text-driven training objectives, prioritizing  textual annotations over image features. This imbalance often leads to modality misalignment, resulting in several limitations: hallucination phenomena \cite{wang2024mllm, wang2023amber}, inadequate comprehension of fine-grained semantic information in images \cite{wu2024towards, lai2024lisa}, and degraded performance in vision-language tasks \cite{lin2024vila}. Existing methods, such as advanced fusion modules \cite{bai2023qwen, ye2023mplug, zhu2023minigpt} or autoregressive tasks integrated into the image branch \cite{tong2024metamorph, fini2024multimodal}, show promise but often introduce significant architectural complexity or require extensive human intervention. These approaches fail to resolve training imbalances, where attention mechanisms favor text tokens over image tokens.


LLava offers a simpler alternative by introducing a lightweight linear layer to bridge the image and text feature spaces, achieving competitive performance and surpassing more complex architectures like Q-Former \cite{li2022blip}. Subsequent enhancements, LLava replaced the linear layer with a multilayer perceptron (MLP), further improving alignment and accuracy. However, recent studies \cite{zhang2024redundancy} reveal that LLava's attention mechanism still assigns insufficient weight to image tokens, limiting its ability to fully exploit visual information. This imbalance exacerbates alignment challenges and reduces the model’s effectiveness in cross-modal tasks.


To address these issues, some approaches have focused on improving image-text alignment by assigning greater importance to image tokens \cite{huang2024opera} or introducing additional encoders \cite{xing2024mitigating}. While these methods mitigate specific alignment issues, they often rely on complex architectural modifications or external supervision signals, increasing computational costs and operational challenges. Thus, achieving balanced and efficient alignment between image and text tokens remains an open problem in MLLMs.

From an information-theoretic perspective, the pre-training objective of MLLMs is to minimize mutual information between input and output textual representations, ensuring accurate text reconstruction. Extending this framework to image-related tasks, we show that L2 loss effectively approximates mutual information, enabling better alignment between image and text features. Building on this insight, we propose a method called Visual Dynamic Embedding-guided pre-training (VDEP). Without altering the architecture of the LLava model, VDEP achieves a significant improvement in image-text alignment.

Specifically, our approach leverages dynamic embeddings generated by the MLP following the visual encoder to supervise the hidden states of image representations. Simultaneously, image tokens are incorporated into the autoregressive training objective, allowing for the joint optimization of text and image during training. This mechanism enables the model to adjust the relative importance of image and text tokens dynamically, effectively balancing their attention distributions.

As illustrated in Figure \ref{lighthouse}, VDEP significantly enhances the model's ability to capture critical visual information, even demonstrating a nuanced understanding of contours within images. Across multiple benchmark datasets, our method delivers state-of-the-art performance and achieves an approximate 5.2\% improvement in RealWorldQA benchmark. Experimental results confirm that the autoregressive latent space alignment strategy mitigates hallucination issues and substantially boosts the model's effectiveness in cross-modal tasks.
To summarize, our key contributions are: 
\begin{itemize}
\item We redefine multimodal alignment as an information recovery process and propose VDEP, which integrates image information restoration into LLava's pre-training process without requiring additional data or architectural modifications.
\item We design a novel hybrid training strategy that dynamically alternates between VDEP mode and the original LLava mode, enhancing alignment performance during pre-training.
\item Experimental results demonstrate that VDEP consistently outperforms the baseline LLava across 13 benchmark datasets, including VQA, MME, and MMB.
\end{itemize}



\section{Related Works}
\label{related_work}

\begin{figure*}[ht]
\vskip 0.2in
\begin{center}
\centerline{\includegraphics[width=1.0\textwidth]{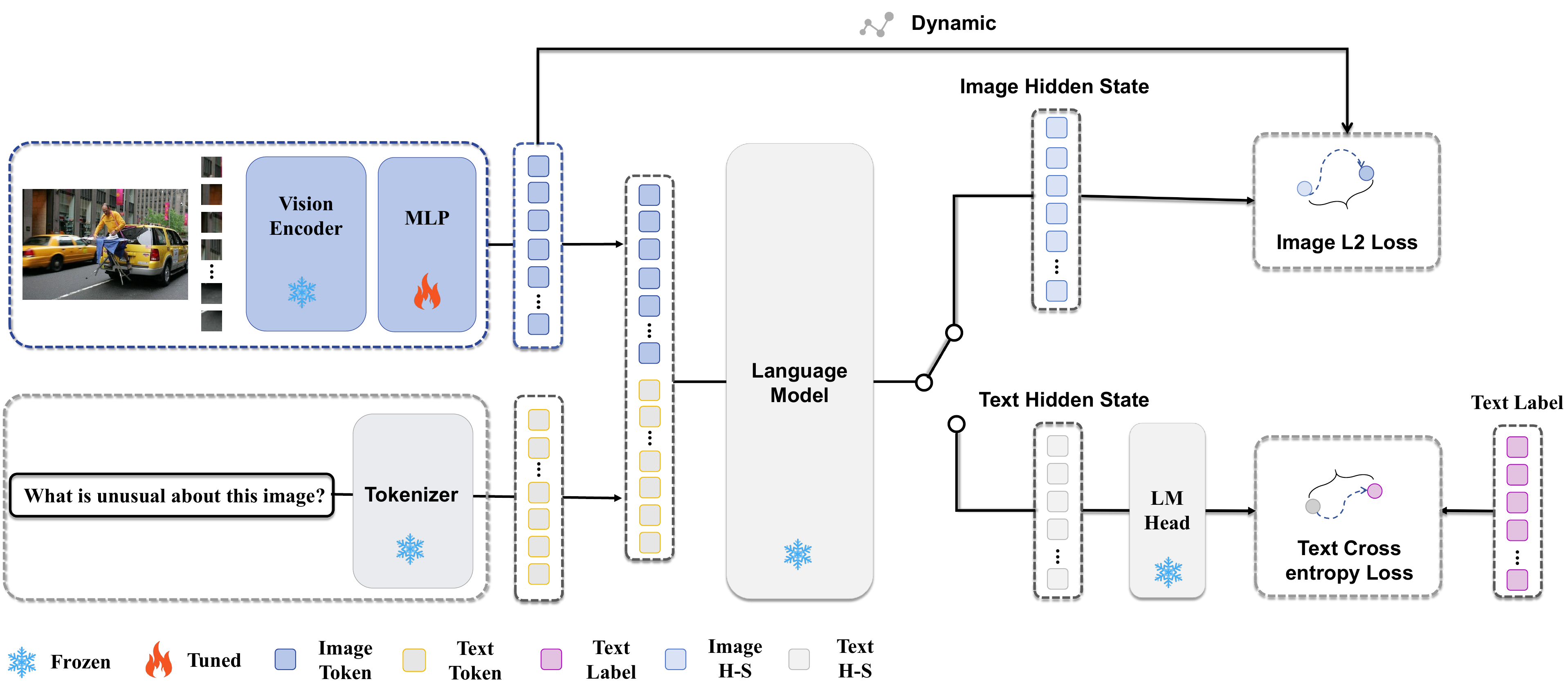}}
\caption{The LLava-VDEP network architecture incorporates two distinct training modes. The VDEP mode performs supervised learning on image data, while the LLava mode is dedicated to supervised learning on text data. During batch training, a ratio parameter is used to control the proportional occurrence of these two modes within each batch, enabling an effective balance in the learning process.}
\label{VRKG}
\end{center}
\vskip -0.2in
\end{figure*}
\subsection{Multimodal Large Language Models}
Language and vision models in single-modal settings inherently face functional limitations, which restrict their potential for further development. In contrast, the rapid emergence of large multimodal models has successfully overcome these limitations, leading to notable improvements in model capabilities and performance. For instance, CLIP\cite{radford2021learning} employs an efficient pre-training strategy based on contrastive learning; however, its similarity computation relies exclusively on basic dot-product operations. In comparison, VILT\cite{kim2021vilt} highlights that traditional models often feature excessively intricate backbone designs for vision and text, whereas the multimodal fusion modules are relatively simplistic. To address this, VILT reallocates computational resources toward the alignment module, thereby improving the effectiveness of multimodal learning. Meanwhile, ALBEF\cite{li2021align} argues that aligning images and text solely through the fusion module presents challenges. To mitigate this, it introduces the ITC (Image-Text Contrastive Learning) module, which performs initial alignment between images and text before passing them to the fusion module. This refinement lays a stronger foundation for subsequent fusion processes. Building on this, BLIP-v2\cite{li2023blip} incorporates the ITC module. It introduces the Q-Former, employing a two-stage pre-training strategy: the first stage leverages a frozen vision encoder to guide multimodal learning. In contrast, the second stage utilizes a frozen text encoder to complete the learning process. However, BLIP-v2 compresses image information by decreasing the number of image tokens, which improves efficiency but introduces limitations in specific downstream tasks. Additionally, LLava\cite{liu2024visual} adopts a more straightforward approach, projecting image features into the text feature space via a linear layer. As the complexity of the projection layer increases, its alignment performance improves significantly.
\begin{figure*}[ht]
\vskip 0.2in
\begin{center}
\centerline{\includegraphics[width=0.9\textwidth]{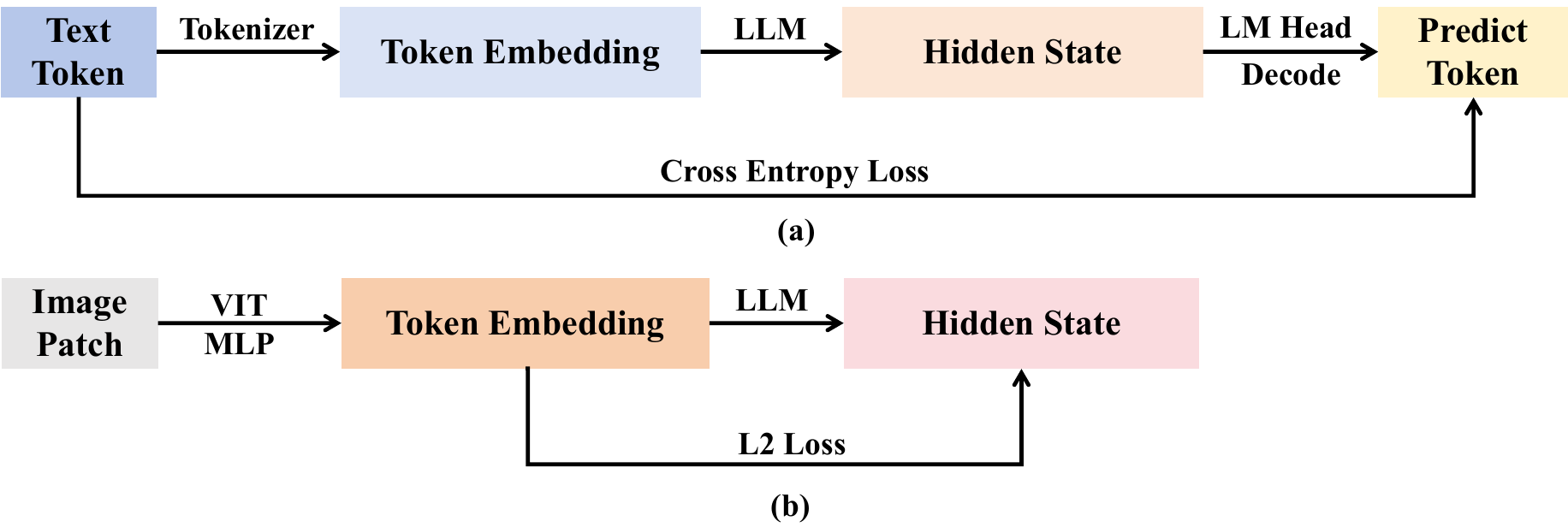}}
\caption{Illustration of our VDEP derivation process.
(a) Text Pre-training: Convert text into embeddings using tokenization. The LLM generates hidden states, which are processed by the LM head to produce predicted tokens. Compute cross-entropy loss with the original input.(b) Image Pre-training: Divide images into patches. Convert patches into embeddings using visual branches without real labels. These embeddings guide the LLM hidden states to reconstruct image information.}
\label{VDEP_derivation}
\end{center}
\vskip -0.2in
\end{figure*}

\subsection{Strategies for Aligning MLLMs}
Research on aligning MLLMs across tasks is divided into two stages: the pre-training stage and the SFT (Supervised Fine-Tuning) stage. 

\textbf{Pre-training stage.} Key strategies include:  
(1) Enhancing semantic information in the visual branch: Text tokens often struggle to capture complete instance-level information from segmented image tokens. To address this, methods such as \cite{zhong2022regionclip, minderer2022simple} use pre-trained region-based models or object detection datasets to extract semantic features. Other approaches \cite{lai2024lisa, yang2023improved, aflalo2024fivl, zhao2023bubogpt} leverage mature models \cite{radford2021learning} to improve cross-modal instance alignment.  
(2) Introducing adapter modules: Adapter modules integrate image and text modalities effectively. For example, \cite{zhang2023llama} uses zero-initialized attention for efficient fine-tuning, while \cite{gao2023llama} decouples instruction-following training from vision-language alignment to reduce interference.  
(3) Improving alignment between feature spaces: Models like LLava \cite{liu2024visual}, Shikra \cite{chen2023shikra}, and MiniGPT \cite{chen2023minigpt} map visual features to text spaces via linear projection adapters. Others, such as QWEN-VL \cite{bai2023qwen} and mPLUG-Owl \cite{ye2023mplug}, employ Perceiver-like architectures, while Q-Former \cite{li2023blip} has become a core adapter in many models \cite{zhu2023minigpt, yu2023reformulating}.  
(4) Incorporating additional tasks: Visual generation tasks \cite{tong2024metamorph} and pixel-level generation objectives \cite{fini2024multimodal} enhance visual understanding and strengthen image-text alignment.  

\textbf{SFT stage.} Key strategies include:  
(5) Developing new fine-tuning instruction sets: Models such as LLava \cite{liu2024visual} and MiniGPT-4 \cite{zhu2023minigpt} integrate image caption data for visually grounded dialogue. MultiInstruct \cite{xu2022multiinstruct} reformulates visual tasks into instruction-tuning formats, while InstructBLIP \cite{NEURIPS2023_9a6a435e} collects data from 11 tasks for instruction fine-tuning. 
M3IT \cite{li2023large} introduces a multimodal multilingual dataset spanning 40 datasets to optimize instruction tuning.

\section{Method}
\label{method}
\subsection{Problem Formulation}


In modality-aligned pre-training, MLLMs aim to establish semantic alignment between image representations ($X_i$) and text representations ($X_t$) by learning their mutual relationships. Specifically, an image $I$ is divided into a series of non-overlapping patches $P = [p_1, p_2, \cdots]$, which are processed using a Vision Transformer (ViT) followed by an MLP projection. The resulting representation forms the input embeddings of the image, denoted as $X_i = [x_{i1}, x_{i2}, \cdots]$. Correspondingly, the textual description of the image is represented as a sequence $X_t = [x_{t1}, x_{t2}, \cdots]$.
Typically, MLLMs employ alignment strategies to generate predicted text $X_t^{\text{p}}$ based on image inputs $X_i$. In an autoregressive training paradigm, this involves treating both $X_i$ and $X_t$ as inputs during training and minimizing the divergence between the predicted text $X_t^{\text{p}}$ and the target text $X_t$. The optimization objective can be expressed as:
\begin{equation}
L_t = - \sum_{t} P(X_t | X_i) \log P(X_t^{\text{p}} | X_i)
\end{equation}
This approach is inspired by the traditional pre-training framework of LLMs, where the model is trained to reconstruct input text by minimizing the difference between the input sequence $X_t$ and the predicted sequence $X_t^{\text{p}}$.
\begin{equation}
L_t = - \sum_{t} P(X_t) \log P(X_t^{\text{p}})
\end{equation}
Within the context of multimodal tasks, MLLMs adopt a training paradigm similar to that of LLMs, prioritizing textual tokens over image data. Consequently, image information is often treated as static prior input, limiting its dynamic integration during the training process. This approach results in an imbalance, where text information dominates over image information during training.
As illustrated in Figure \ref{VDEP_derivation}.a, this inconsistency can be better understood through the lens of information theory. From an information-theoretic perspective, the goal of an autoregressive model is to maximize the mutual information $I(X_t;X_t^{\text{p}})$ between the input text $X_t$ and the predicted text $X_t^{\text{p}}$. This ensures that the model effectively captures and reconstructs the semantic content of the input text. This objective can be formulated as:
\begin{equation} 
\textstyle
I(X_t; X_t^{\text{p}}) = H(X_t) - H(X_t | X_t^{\text
{p}})
\label{LLM_mutaul_info_equ}
\end{equation}
During training, maximizing $I(X_t; X_t^{\text{p}})$ is equivalent to minimizing the conditional entropy $H(X_t | X_t^{\text{p}})$
. This relationship implies that as mutual information increases:
\begin{equation}
\begin{aligned}
H(X_t \mid X_t^{\text{p}}) &\to 0 \\
I(X_t; X_t^{\text{p}}) &\to H(X_t)
\end{aligned}
\end{equation}
Since the cross-entropy loss is inherently tied to $H(X_t | X_t^{\text{p}})$, minimizing the cross-entropy loss effectively reduces uncertainty, enhancing the model’s ability to reconstruct the input text $X_t$ from its predictions $X_t^{\text{p}}$. Analysis of the equation \ref{LLM_mutaul_info_equ} reveals that the visual information $X_i$ is not explicitly integrated into the optimization objective. This discrepancy in multimodal alignment between visual and textual representations leads to three critical challenges:
(1) The feature distribution of image information $X_i$ is not directly optimized within the training objective, which may limit the model's ability to fully leverage visual inputs;
(2) The optimization objective for text generation $X_t^{\text{p}}$ is primarily focused on reconstructing textual information $X_t$
, neglecting the need for effective cross-modal information fusion.
From an information-theoretic perspective, this inconsistency arises because the model fails to fully exploit the image input $X_i$ to maximize the mutual information $I(X_i; X_t^{\text{p}})$
, thereby limiting the overall effectiveness of multimodal alignment.

\subsection{Dynamic Vision Autoregressive Training}

As shown in Figure \ref{VDEP_derivation}, text data includes instance-level token labels, allowing the use of cross-entropy loss to quantify discrepancies between reconstructed outputs and input tokens. In contrast, image data lacks explicit instance-level labels, making the direct application of cross-entropy loss computationally prohibitive. To address this limitation, we propose a redesigned autoregressive training framework that incorporates image representations into the optimization objective:
\begin{equation}
\textstyle
I(X_I; X^{h}_I) = H(X_I) - H(X_I|X^{h}_I)
\end{equation}
Here, $X^{h}_I$ represents the hidden vector generated by the LLM at the corresponding position. This process is depicted in Figure \ref{VDEP_derivation}.b. In the absence of explicit token-level labels for images, we use embedding vectors to reconstruct the input image representations. This approach aligns the optimization process for image data with the token-level objectives used for text data. 
To maximize mutual information, we minimize the difference between $X_I$ and $X^{h}_I$. Considering that cross-entropy is not well-suited for the measurement of similarity between vectors, this study adopts the L2 norm to quantify the Euclidean distance-based difference between two vectors.
\begin{equation}
\textstyle
\mathcal{L}_i = \left\lVert X_I - X^{h}_I \right\rVert_2^2
\end{equation}
By minimizing the loss function $\mathcal{L}_i$, the hidden vector $X^{h}_I$ converges toward the original image embedding vector $X_I$, as described by:
\begin{equation}
\textstyle
\lim_{\mathcal{L}_i \to 0} X^{h}_I = X_I
\end{equation}
When this condition is satisfied, the conditional entropy $H(X_I | X^{h}_I)$ approaches zero, and the mutual information $I(X_I; X^{h}_I)$ becomes equal to the entropy $H(X_I)$.
\begin{equation}
\begin{aligned}
\textstyle
\lim_{X^{h}_I \to X_I} I(X_I; X^{h}_I) &= H(X_I)
\\
\lim_{X^{h}_I \to X_I} H(X_I | X^{h}_I) &= 0
\end{aligned}
\end{equation}
Therefore, the optimization objective is defined as:
\begin{equation}
\lim_{\mathcal{L}_i \to 0} \Big( X^{h}_I \to X_I, \, \, H(X_I | X^{h}_I) \to 0 \Big)
\end{equation}
This result shows that the hidden vector $X^{h}_I$ successfully captures and reconstructs the semantic information encoded in the original image embedding vector $X_I$.

\subsection{Hybrid Multimodal Alignment Training}
According to the above analysis, we introduce a new objective function, $\mathcal{L}$, which balances the contributions of both visual and textual modalities:
\begin{equation}
\begin{aligned}
\textstyle
\mathcal{L} =\mathcal{L}_t +  \alpha \mathcal{L}_i
\end{aligned}
\end{equation}
where $\mathcal{L}_t$ and $\mathcal{L}_v$ denote the loss functions for text and image modalities, respectively. The parameter $\alpha \in [0, 1]$ controls the relative importance of vision modality. By tuning $\alpha$, we can dynamically adjust the balance between modalities to optimize overall model performance.
The training process of LLava can be divided into two main stages: the pre-training stage and the supervised fine-tuning stage. As shown in Figure \ref{VRKG}, during the pre-training stage, we employed a hybrid multimodal alignment training strategy, combining the VDEP mode and the LLava mode. In the supervised fine-tuning stage, we consistently followed the original LLava framework.
In the pre-training stage, the process prioritizes optimizing the text space to align the image space with the text space. In MLLMs, the number of text tokens is significantly smaller than the number of image tokens. Training both modalities jointly without distinction can cause the model to overfit background information in images, hindering its ability to match the performance of text-only supervised training. Furthermore, it can introduce noise into the training process, potentially disrupting the alignment between image and text modalities.
To address this, we utilize the distribution derived from text data to align the image space with the text space. This requires the text distribution to first reach a stable state. Our mixed multimodal alignment strategy, which alternates the optimization of image and text losses during each forward and backward propagation, ensures balanced learning across modalities. Specifically, within each batch, the data is evenly divided, with one half (randomly selected) used to compute the image loss and the other half to compute the text loss. This
\begin
{table*}[t]
\caption
{Comparison of VDEP (Ours) and LLava series Across visual question-answering Datasets with Different Model Sizes.}
\label{main_vqa}
\vskip
 0.15in
\begin
{center}
\begin
{small}
\begin
{sc}
\scalebox
{1.0}{
\begin
{tabular}{lcccccc}
\toprule
\textbf{METHOD} & \textbf{VQA\textsuperscript{ok}} & \textbf{GQA} & \textbf{VIZWIZ} & \textbf{VQA\textsuperscript{T}} & \textbf{RWQA} & \textbf{SQA\textsuperscript{I}} \\
\midrule
\textbf{\emph{TinyLLava-3B}} & & & & & & \\ 
TinyLLava & \multicolumn{1}{c}{56.71} & \multicolumn{1}{c}{61.07} & \multicolumn{1}{c}{34.30} & \multicolumn{1}{c}{50.88} & \multicolumn{1}{c}{53.99} & \multicolumn{1}{c}{71.24} \\
\rowcolor[HTML]{ededed} 
TinyLLava-VDEP (ours) & \multicolumn{1}{c}{\textbf{57.97}} & \multicolumn{1}{c}{\textbf{61.67}} & \multicolumn{1}{c}{\textbf{37.58}} & \multicolumn{1}{c}{\textbf{51.73}} & \multicolumn{1}{c}{\textbf{54.25}} & \multicolumn{1}{c}{\textbf{71.39}} \\
Change & \multicolumn{1}{c}{\textcolor{darkgreen}{+1.26}} & \multicolumn{1}{c}{\textcolor{darkgreen}{+0.60}} & \multicolumn{1}{c}{\textcolor{darkgreen}{+3.28}} & \multicolumn{1}{c}{\textcolor{darkgreen}{+0.85}} & \multicolumn{1}{c}{\textcolor{darkgreen}{+0.26}} & \multicolumn{1}{c}{\textcolor{darkgreen}{+0.15}} \\
\midrule
\textbf{\emph{LLava-v1.5-7B}} & & & & & & \\ 
LLava & \multicolumn{1}{c}{54.32} & \multicolumn{1}{c}{62.12} & \multicolumn{1}{c}{50.34} & \multicolumn{1}{c}{46.16} & \multicolumn{1}{c}{54.80} & \multicolumn{1}{c}{66.80} \\
\rowcolor[HTML]{ededed} 
LLava-VDEP (Ours) & \multicolumn{1}{c}{\textbf{57.68}} & \multicolumn{1}{c}{\textbf{62.50}} & \multicolumn{1}{c}{\textbf{50.37}} & \multicolumn{1}{c}{\textbf{46.76}} & \multicolumn{1}{c}{\textbf{57.64}} & \multicolumn{1}{c}{\textbf{68.36}} \\
Change & \multicolumn{1}{c}{\textcolor{darkgreen}{+3.36}} & \multicolumn{1}{c}{\textcolor{darkgreen}{+0.38}} & \multicolumn{1}{c}{\textcolor{darkgreen}{+0.03}} & \multicolumn{1}{c}{\textcolor{darkgreen}{+0.60}} & \multicolumn{1}{c}{\textcolor{darkgreen}{+2.84}} & \multicolumn{1}{c}{\textcolor{darkgreen}{+1.56}} \\
\bottomrule
\end
{tabular}
}
\end
{sc}
\end
{small}
\end
{center}
\vskip
 -0.1in
\end
{table*}
\begin
{table*}[t]
\caption
{Comparison of VDEP (Ours) and LLava series on benchmarks for instruction-following LMMs with Different Model Sizes.}
\label{MLLM_Benchmark}
\vskip
 0.15in
\begin
{center}
\begin
{small}
\begin
{sc}
\scalebox
{0.9}{
\begin
{tabular}{@{}cccccccccc@{}}
\toprule
\multirow{2}{*}{\textbf{METHOD}} & \multirow{2}{*}{\textbf{POPE}} & \multirow{2}{*}{\textbf{SEEDB\textsuperscript{I}}} & \multirow{2}{*}{\textbf{AI2D}} & \multirow{2}{*}{\textbf{MM-Vet}} & \multirow{2}{*}{\textbf{MMMU}} & \multirow{2}{*}{\textbf{MMTB}} & \multirow{2}{*}{\textbf{OCRB}} & \multicolumn{2}{c}{\textbf{MMBench}} \\ 
\cmidrule
(lr){9-10}
& & & & & & & & \textbf{en} & \textbf{cn} \\ 
\midrule
\textbf{\emph{TinyLLava-3B}} & & & & & & & & & \\ 
TinyLLava & 85.93 & 68.54 & 59.75 & 33.00 & 33.80 & 48.93 & \textbf{345} & \textbf{67.88} & \textbf{45.07} \\
\rowcolor
[HTML]{EDEDED}
TinyLLava-VDEP (Ours) & \textbf{86.98} & \textbf{69.35} & \textbf{60.85} & \textbf{36.00} & \textbf{33.80} & \textbf{49.08} & 343 & 66.70 & 41.87 \\
Change & \textcolor{darkgreen}{+1.05} & \textcolor{darkgreen}{+0.81} & \textcolor{darkgreen}{+1.10} & \textcolor{darkgreen}{+3.00} & \textcolor{darkgreen}{+0.00} & \textcolor{darkgreen}{+0.15} & -2.00 & -1.18 & -3.20 \\
\midrule
\textbf{\emph{LLava-v1.5-7B}} & & & & & & & & & \\ 
LLava & 85.85 & 66.10 & 55.63 & 31.10 & 31.20 & 47.94 & 297 & 64.97 & 57.90 \\
\rowcolor
[HTML]{EDEDED}
LLava-VDEP (Ours) & \textbf{86.20} & \textbf{66.70} & \textbf{56.57} & 30.60 & 30.80 & \textbf{48.00} & \textbf{326} & \textbf{66.81} & \textbf{58.23} \\
Change & \textcolor{darkgreen}{+0.35} & \textcolor{darkgreen}{+0.60} & \textcolor{darkgreen}{+0.94} & -0.50 & -0.40 & \textcolor{darkgreen}{+0.06} & \textcolor{darkgreen}{+29} & \textcolor{darkgreen}{+1.84} & \textcolor{darkgreen}{+0.33} \\
\bottomrule
\end
{tabular}
}
\end
{sc}
\end
{small}
\end
{center}
\vskip
 -0.1in
\end
{table*}
approach preserves the inherent characteristics of the text distribution while using it to align the image space.

\section{Experiments}
\label{experments}

\textbf{Datasets.}
The pre-training and fine-tuning datasets used in this work are identical to those utilized in LLava-v1.5.
For pre-training, we use a subset of the LAION/CC/SBU dataset filtered for balanced concept coverage and enriched with BLIP-generated captions. 
For instruction tuning, we use a combination of COCO\cite{lin2014microsoft}, GQA\cite{hudson2019gqa}, OCR-VQA\cite{mishra2019ocr}, TextVQA\cite{singh2019towards}, and VisualGenome\cite{krishna2017visual} datasets. The details of the datasets are in the Appendix. 

\textbf{Tasks and Evaluation.}
Building upon previous studies, we conduct experiments on a range of visual question-answering benchmarks, including OK-VQA\cite{marino2019ok}, GQA\cite{hudson2019gqa}, VizWizQA\cite{Bigham_Jayant_Ji_Little_Miller_Miller_Tatarowicz_White_White_Yeh_2010}, TextVQA\cite{singh2019towards}, RealWorldQA\cite{grok15v}, and ScienceQA\cite{lu2022learn}. Additionally, we evaluate our method on widely recognized general multimodal benchmarks, such as MMBench\cite{liu2025mmbench}, POPE\cite{li2023evaluating}, SEED\cite{li2023seed}, AI2D\cite{kembhavi2016diagram}, MM-Vet\cite{yu2023mm}, MMMU\cite{yue2024mmmu}, MMTB\cite{ying2024mmt}, OCR-VQA\cite{mishra2019ocr}, and MME\cite{fu2024mmecomprehensiveevaluationbenchmark}. We utilize the evaluation framework lmms-eval\cite{zhang2024lmmsevalrealitycheckevaluation, lmms_eval2024}, as recommended by LLava, which integrates the evaluation capabilities of various benchmarks.

\textbf{Models.} To comprehensively verify the effectiveness of the VDEP, we employ base models of different parameter scales. Specifically, we utilize TinyLLava~\cite{zhou2024TinyLLava} and LLava-v1.5~\cite{liu2024improved} as our base models, whose model sizes are 3B and 7B, respectively. A series of carefully designed experiments are conducted to evaluate their performance.

\textbf{Baseline and Implementation.} 
We compare TinyLLava-VDEP with TinyLLava and LLava-v1.5-VDEP with LLava-v1.5. We thoroughly tune the hyperparameters for different-scale base models and report the best performance. The details of different-scale base models can be found in the Appendix. During the pre-training, we employ two training strategies, VDEP and LLava. We double the input data to ensure these two strategies receive fair and sufficient training. 
To enable smooth transitions between the two strategies, we introduce a novel special token, \textless auto\_image\textgreater. Specifically, when image data undergoes autoregressive training, the \textless auto\_image\textgreater  token is used instead of the conventional \textless image\textgreater  token. 

\section{Empirical Results and Analysis}
We evaluate the visual capabilities of models trained with VDEP across various visual question-answering tasks and
\begin
{table*}[ht]
\caption
{Comparison of LLava-VDEP (Ours) and LLava-v1.5 on MME Tasks with Different Model Sizes.}
\label{MME_benchmark}
\vskip
 0.15in
\begin
{center}
\begin
{small}
\begin
{sc}
\scalebox
{0.9}{
\begin
{tabular}{@{}p{4.0cm}ccccccc@{}}
\toprule
\multirow{2}{*}{\raggedright\textbf{METHOD}} & \multirow{2}{*}{\textbf{Perception}}  & \multicolumn{1}{c}{\textbf{Commonsense QA}} & \multicolumn{4}{c}{\textbf{Coarse-grained Perception Tasks}} & \textbf{Total} \\
 \cmidrule
(lr){4-7}
& & \textbf{(Reasoning)} & \textbf{Existence} & \textbf{Count} & \textbf{Position} & \textbf{Color} & \textbf{Scores} \\ 
\midrule
\textbf{\emph{TinyLLava-3B}} & & & & & & & \\ 
TinyLLava & 1488.30 & 120.71 & 185.00 & 143.33 & 133.33 & 180.00 & 762.37 \\
\rowcolor[HTML]{ededed} 
TinyLLava-VDEP (Ours) & \textbf{1499.08} & \textbf{130.70} & \textbf{200.00} & \textbf{158.33} & \textbf{138.33} & \textbf{180.00} & \textbf{807.36} \\
Change & \textcolor{darkgreen}{+10.78} & \textcolor{darkgreen}{+9.99} & \textcolor{darkgreen}{+15.00} & \textcolor{darkgreen}{+15.00} & \textcolor{darkgreen}{+5.00} & \textcolor{darkgreen}{+0.00} & \textcolor{darkgreen}{+44.99} \\
\midrule
\textbf{\emph{LLava-v1.5-7B}} & & & & & & & \\ 
LLava & 1510.72 & 135.71 & 190.00 & \textbf{158.33} & 128.33 & 175.00 & 787.37 \\
\rowcolor[HTML]{ededed} 
LLava-VDEP (Ours) & \textbf{1516.60} & \textbf{136.00} & \textbf{190.00} & 153.30 & \textbf{135.00} & \textbf{180.00} & \textbf{794.30} \\
Change & \textcolor{darkgreen}{+5.88} & \textcolor{darkgreen}{+0.29} & \textcolor{darkgreen}{+0.00} & -5.03 & \textcolor{darkgreen}{+6.67} & \textcolor{darkgreen}{+5.00} & \textcolor{darkgreen}{+6.93} \\
\bottomrule
\end
{tabular}
}
\end
{sc}
\end
{small}
\end
{center}
\vskip
 -0.1in
\end
{table*}
\begin
{table}[t]
\caption{Ablation study on the hyperparameter $\alpha$
, which represents the variation of the image loss weight.}
\label{alpha}
\vskip
 0.15in
\begin
{center}
\begin
{small}
\begin
{sc}
\scalebox
{1.0}{
\begin
{tabular}{lcccc}
\toprule
\textbf{$\alpha$} & \textbf{RWQA} & \textbf{MME\textsuperscript{P}} & \textbf{MMB} & \textbf{VQA\textsuperscript{ok}} \\
\midrule
\textbf{\emph{LLava-VDEP-7B}} \\
\emph{w}/ 0.1 & 55.42 & 1479.00 & 62.25 & 57.33 \\
\emph{w}/ 0.01 & 56.73 & 1504.99 & 62.52 & 55.70 \\
\midrule
\rowcolor
[HTML]{ededed}
\emph{w}/ 0.001  & \textbf{57.64} & \textbf{1515.60} & \textbf{62.52} & \textbf{57.68} \\
\bottomrule
\end
{tabular}}
\end
{sc}
\end
{small}
\end
{center}
\vskip
 -0.1in
\end
{table}

general multimodal benchmarks. This novel reconstruction of image information achieves highly competitive performance across different model scales. Our proposed method consistently demonstrates state-of-the-art performance on most evaluation metrics, outperforming baselines. Unless stated otherwise in the following experiments, the image loss weight $\alpha$ is set to 0.001, and the data ratio is fixed at 1.0.

\subsection{Performance Comparison On Benchmarks}

To evaluate the effectiveness of our model in general visual question-answering tasks, we perform comprehensive evaluations on a diverse set of advanced benchmarks. The results, summarized in Table \ref{main_vqa}, demonstrate that the VDEP series consistently outperforms the baseline across all benchmarks.
On the OK-VQA benchmark, which requires external knowledge for answering VQA problems, the VDEP 7B model improves by 6.18\%. On the VizWiz dataset, characterized by low-quality and blurred images, the VDEP 3B model outperforms the baseline by 9.56\%. On the RealWorldQA benchmark, which assesses spatial reasoning in realworld scenarios, the VDEP 7B model surpasses the baseline by 5.18\%. On the large-scale multimodal scientific question-answering task, the VDEP 7B model achieves a 2.33\% improvement over the baseline.

To further investigate VDEP's capabilities in visual perception, we conduct experiments on the MME benchmark, which evaluates both perceptual and cognitive abilities. The results, summarized in Table \ref{MME_benchmark}, show notable improvements in visual perception, with the 3B model achieving substantial gains across all metrics. We also evaluate the VDEP
\begin
{table}[t]
\caption
{Ablation study on the hyperparameter Data Ratio, which represents the proportion of different VDEP and LLava patterns in the pre-training stage.}
\label{data_ratio}
\vskip
 0.15in
\begin
{center}
\begin
{small}
\begin
{sc}
\scalebox
{0.9}{
\begin
{tabular}{lcccc}
\toprule
\textbf{Data Ratio} & \textbf{RWQA} & \textbf{MME\textsuperscript{P}} & \textbf{MMB} & \textbf{VQA\textsuperscript{ok}} \\
\midrule
\textbf{\emph{LLava-VDEP-7B}} \\
\emph{w}/ 0.5 & 54.25 & 1439.16 & 58.90 & 55.80 \\
\emph{w}/ 0.8 & 57.12 & 1509.47 & 59.36 & 57.26 \\
\midrule
\rowcolor
[HTML]{ededed}
\emph{w}/ 1.0 & \textbf{57.64} & \textbf{1515.60} & \textbf{62.52} & \textbf{57.68} \\
\bottomrule
\end
{tabular}
}
\end
{sc}
\end
{small}
\end
{center}
\vskip
 -0.1in
\end
{table}
series on mainstream multimodal benchmarks to ensure a fair comparison and achieve competitive results.
On the MM-Vet benchmark, which evaluates the integration of visual and language capabilities across 16 complex multimodal tasks, the VDEP 3B model improves by 9.09\% over the baseline. On the MMT-Bench benchmark, which assesses advanced reasoning and instruction-following abilities across 32 meta-tasks and 162 sub-tasks, the VDEP series consistently outperforms the baseline, demonstrating its strengths in visual recognition, localization, reasoning, and planning. On the MMBench benchmark, which measures 20 fine-grained multimodal abilities, the VDEP series delivers a performance that matches or surpasses the state-of-the-art.
Lastly, on the AI2D benchmark, a dataset focused on multiple-choice questions involving scientific diagrams with textual content, the VDEP model achieves leading performance, highlighting its strong ability to understand textual information within images.

\subsection{Further Analysis}
MLLMs are trained in two stages. During the pre-training stage, the objective is to ensure that the semantic representations of different modalities, such as text and images, are aligned in a shared latent space. In the Supervised Fine-Tuning (SFT) stage, text tokens are leveraged to query image tokens, enabling tasks that require instance-level semantic understanding. Figure \ref{lighthouse} illustrates how VDEP produces more concentrated attention maps than baseline models, with clear contours outlining core objects. This visualization highlights VDEP's ability to focus on key visual elements, aiding problem-solving. (1) \textbf{Denoising.} This advantage is particularly evident in the VizWiz dataset, where low-quality images and key instance objects are off-center. VDEP's attention mechanism effectively suppresses noise in low-quality images and focuses on critical visual elements, resulting in substantial performance improvements.
(2) \textbf{Location.} VDEP consistently focuses on key regions for multi-turn question-answering tasks, enhancing its contextual understanding. In the MME task, VDEP enhances the ability to localize key instances within images, demonstrating that text tokens can precisely correspond to critical positions in the visual input. These results highlight the strong modality alignment capabilities of VDEP during the pre-training stage. (3) \textbf{Capture.} In VQA task, VDEP's attention maps accurately capture object boundaries and spatial relationships, providing superior visual features for scene-based question-answering. 

However, VDEP performs poorly on specific mainstream multimodal benchmarks, such as MMBench, despite these strengths, when using the 3B model. Specifically, the 3B model performs waaker in Chinese and English scenarios. This performance gap may be attributed to an over-reliance on visual supervision, potentially limiting the model's ability to process complex linguistic inputs. In contrast, the 7B model achieves notable improvements, indicating a better balance between visual and language features. This ultimately leads to enhanced performance on both Chinese and English tests. In the ablation study, we analyze the impact of varying image loss hyperparameters across different model scales, identifying the optimal range for balancing visual and language feature integration.


\subsection{Ablation Study}
\textbf{Hyperparameters $\mathbf{\alpha}$.} As illustrated in Table \ref{alpha}, with decreasing hyperparameter $\alpha$, the overall performance of the model exhibits a consistent improvement in performance metrics across multiple benchmarks. This observation suggests that reducing the weight assigned to the image loss notably improves the model's performance. 
The underlying reason for this phenomenon lies in the disparity between the number of image tokens and text tokens, with the former being significantly larger. This imbalance often leads to a higher proportion of background tokens in image data. When $\alpha$ is relatively large, the model tends to overfit these background tokens, i.e., the model disproportionately focuses on less informative regions of the image, thereby introducing noise that impairs the effectiveness of text alignment during training. 
By contrast, a smaller $\alpha$ alleviates the constraints of image reconstruction, reducing the influence of background noise and enabling more effective text-image alignment, thereby promoting superior performance in multimodal tasks.

\textbf{Hyperparameters Data Ratio.} As shown in Table \ref{data_ratio}, we utilize the VDEP framework to train the model with varying text-to-image data ratios and assess its performance across multiple multimodal benchmarks. By adjusting the ratio of VDEP mode to LLava mode within a batch during pre-training, we control the proportion of image reconstruction data, where a higher ratio indicates a greater amount of image data.
The results in Table \ref{data_ratio} demonstrate a clear trend: as the proportion of image data increases, the model's overall performance improves consistently across multiple test datasets. This phenomenon is attributed to the greater challenge of simultaneously optimizing regression tasks for both images and text, as it requires balancing competing objectives compared to optimizing only the text regression task. A lack of sufficient image data during pre-training leads to suboptimal learning of all tasks, resulting in weaker alignment between modalities and ultimately degrading the model’s overall performance.

\section{Conclusion}
\label{conclusion}
In this paper, we revisit the optimization objective of MLLMs through the lens of information theory. Existing MLLMs, shaped by the autoregressive paradigm of LLMs, prioritize text generation at the expense of effectively leveraging visual information. To address this limitation, we introduce a novel approach, VDEP, which directly incorporates image reconstruction into the optimization objective of MLLMs. Our proposed VDEP framework seamlessly integrates image reconstruction into standard models without requiring any architectural modifications. Extensive experiments conducted on 13 benchmark datasets demonstrate VDEP's ability to address these challenges and significantly improve visual content comprehension.

\section*{Impact Statement}
In this paper, we propose a novel paradigm for multimodal alignment, named Vision Dynamic Embedding-Guided Pre-training. Grounded in information theory, this approach incorporates the image reconstruction task as an explicit component of the autoregressive objectives in multimodal large models. This paradigm offers a streamlined and effective framework for aligning MLLMs, emphasizing the critical role and efficacy of image reconstruction in facilitating image-text alignment. The experimental setup and data processing in our study adhere to the principles outlined by the LLava dataset.

\nocite{langley00}

\bibliographystyle{icml2025}

\newpage
\appendix
\onecolumn
\begin{figure*}[!t]
  \centering
  \includegraphics[width=1.0\textwidth]{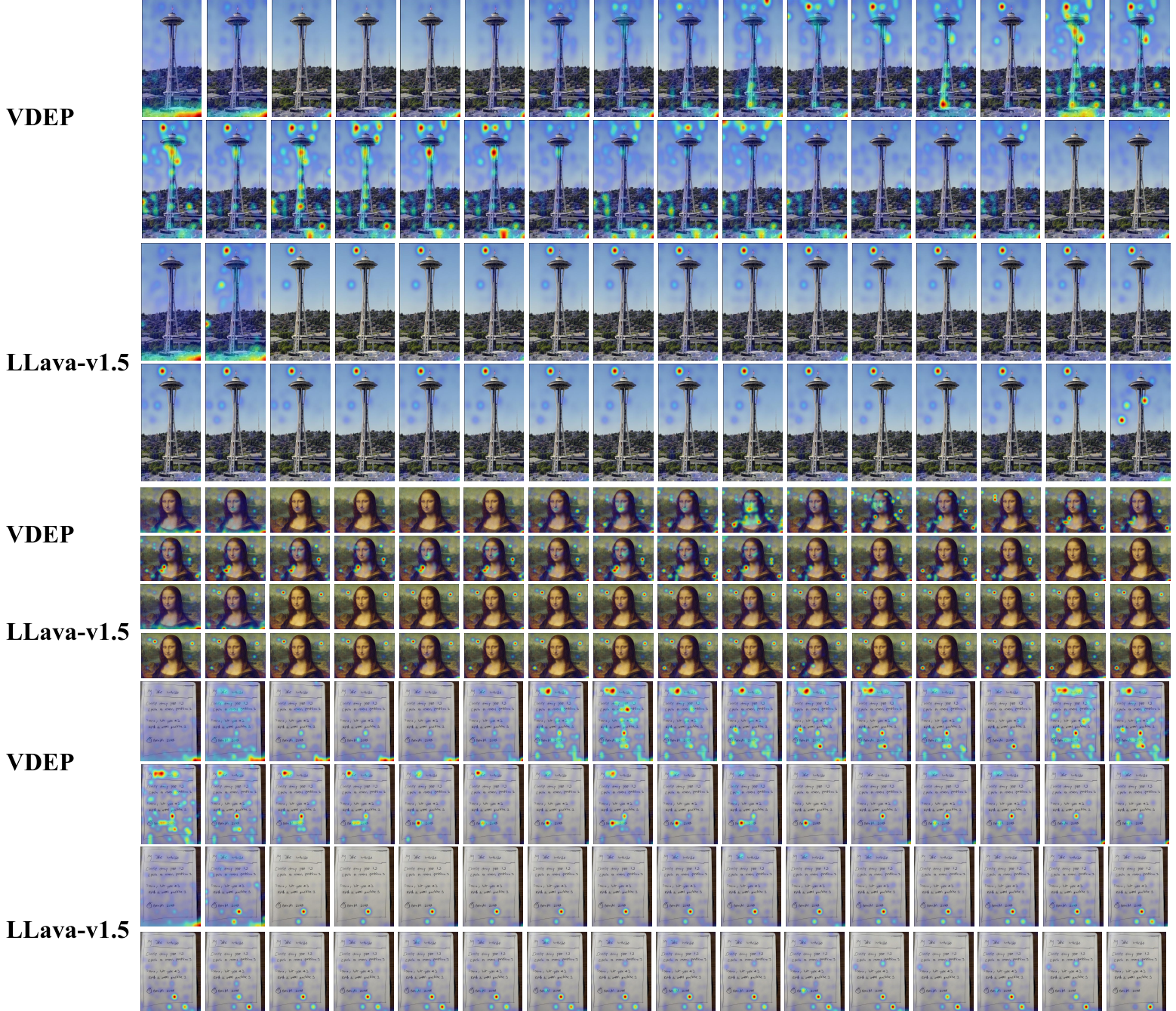}
  \caption{Layer-wise attention visualization of visual-to-instruction information flow. Displayed from top to bottom
are the attention heatmaps from LLava-v1.5-7B and LLava-v1.5-7B-VDEP, respectively. The
example is derived from LLava-Bench (Liu et al., 2024b) and the query is "Describe this photo in detail".
}
  \label{attenion_map}
\end{figure*}

\section{Implementation Details.}

We design a series of experiments to rigorously evaluate the effectiveness of our proposed method across models of varying scales. These experiments involve two models with different parameter sizes: 3B, 7B.
For the 3B model, we use the TinyLLava architecture in our experiments. Within this framework, SigLIP is the visual encoder, while Phi-2 is the language model. For the 7B model, we use the pre-trained CLIP ViT-L/14 ($336^2$) as the visual encoder, combined with the Vicuna v1.5 language model for experiments.
pre-training is conducted on the CC-558K dataset with a $1\times10^{-3}$ learning rate. After pre-training, fine-tuning is performed on the mix-665K dataset with a learning rate of $2\times10^{-5}$. All experiments are conducted on a hardware system with eight NVIDIA A100 GPUs, each with 40GB of memory, to meet the computational requirements. In addition, detailed training steps and specific rules of the implementation plan are fully presented in the appendix.
Our training strategy employs a mixed autoregressive pre-training approach with a strict 1:1 ratio of image data to text data. The image data is sourced from the CC-558K dataset, as in pre-training.. During the SFT stage, our experimental settings match the LLava models.

\section{BenchMarks.}
\subsection{Visual Question Answering}
We conduct experiments on visual question-answering benchmarks, including, OK-VQA, GQA, VizWizQA, TextVQA, RealWorldQA, and ScienceQA. 
OK-VQA includes questions that necessitate external knowledge beyond the multimodal inputs provided. 
GQA is specifically designed to assess the reasoning capabilities of the model. 
VizWizQA is composed of question-answer pairs derived from visually impaired users. 
TextVQA places a greater emphasis on evaluating the model's ability to comprehend text within natural scenes. 
RealWorldQA is a benchmark specifically designed to evaluate the spatial understanding capabilities of multimodal AI models in real-world contexts. 
ScienceQA comprises multimodal multiple-choice questions across a diverse range of science topics. 
These datasets are strategically selected to evaluate our method's capacity to understand comprehensively and reason across diverse visual contexts and knowledge domains.

\textbf{OK-VQA:} OK-VQA(Outside Knowledge VQA)\cite{marino2019ok} is a visual question answering dataset that requires external knowledge. The answers to the questions cannot be inferred solely from the image but also need to incorporate common sense or world knowledge. This dataset evaluates the model's ability in the intersection of vision and knowledge reasoning.  

\textbf{GQA:} GQA(Graph Question Answering)\cite{hudson2019gqa} generates questions and answers based on image scene graphs, focusing on structured reasoning. It emphasizes logical analysis and challenges the model's depth of understanding of semantics and context. 

\textbf{VizWizQA:} VizWizQA\cite{Bigham_Jayant_Ji_Little_Miller_Miller_Tatarowicz_White_White_Yeh_2010} is designed for visually impaired users, with questions originating from real user requests. The images exhibit strong diversity in quality and content. This dataset includes more noise and ambiguous information, making it suitable for evaluating models in real-world application scenarios.  

\textbf{TextVQA:} TextVQA\cite{singh2019towards} focuses on textual information in images, requiring models to recognize and comprehend text within images to answer questions. It drives research on the integration of visual and textual information, expanding the boundaries of visual question answering.

\textbf{RealWorldQA:} RealWorldQA\cite{grok15v} features images and questions sourced from real-world scenarios, encompassing diverse content from daily life. The dataset imposes higher requirements on the model's generalization ability and adaptability to complex scenes.

\textbf{ScienceQA:} ScienceQA\cite{lu2022learn} is a multimodal question answering dataset combining images and scientific questions, covering multiple scientific topics such as physics and biology. It bridges AI technology with the field of science education, promoting intelligent question answering applications in educational contexts.

\subsection{General Multimodal Benchmarks}
We evaluate our proposed method on general multimodal benchmarks, including MME, MMBench, SEED-Bench, POPE, AI2D, MM-Vet, MMMU, and MMT-Bench. 
MME measures both perception and cognition abilities on a total of 14 subtasks. 
MMBench comprehensively evaluates a model's multimodal capabilities in Chinese and English contexts. 
SEED-Bench focuses on assessing generative comprehension in multimodal large language models. POPE evaluates the extent of multimodal hallucinations present in a model. 
AI2D assesses a model's ability to interpret scientific diagram inputs. MM-Vet evaluates the multimodal conversational skills of a model using GPT-4 as a benchmark. 
MMMU is designed to assess multimodal models on extensive multi-disciplinary tasks that require college-level subject knowledge and deliberate reasoning. 
MMT-Bench is a comprehensive benchmark developed to evaluate MLLMs across a wide range of multimodal tasks requiring expert knowledge and deliberate visual recognition, localization, reasoning, and planning.
These diverse benchmarks provide a comprehensive framework for evaluating the performance and capabilities of our proposed method in multimodal learning.

\textbf{MME:} MME\cite{fu2024mmecomprehensiveevaluationbenchmark}, short for Multimodal Evaluation
, is a comprehensive multimodal benchmark designed to evaluate the ability of models to understand and process information across multiple modalities, including vision, text, and audio. It provides a standardized framework to measure performance on tasks requiring cross-modal reasoning and understanding, making it an essential tool for assessing the generalization of multimodal large language models (MLLMs).

\textbf{MMBench} MMBench(Multimodal Benchmark)\cite{liu2025mmbench} is a task-driven benchmark that focuses on systematically evaluating multimodal models across diverse real-world application scenarios, such as visual question answering, image captioning, and video understanding. Its emphasis on practical use cases highlights its importance for assessing the practical utility of MLLMs.

\textbf{SEED:} SEED(Spatial and Entity-aware Evaluation Dataset)\cite{li2023seed} is a benchmark specifically designed to evaluate the spatial and entity reasoning capabilities of multimodal models. By incorporating complex spatial relationships and entity-based queries, SEED tests a model’s ability to perform fine-grained reasoning, which is critical for tasks such as scene understanding and object-oriented question answering.

\textbf{POPE:} POPE(Perceptual and Object-aware Performance Evaluation)\cite{li2023evaluating} focuses on evaluating the perceptual understanding and object-centric reasoning of multimodal models. It emphasizes tasks like object detection, recognition, and spatial awareness, making it a key benchmark for assessing models' performance in visually grounded tasks.

\textbf{AI2D:} AI2D(Allen Institute for AI Diagram Dataset)\cite{kembhavi2016diagram} is a dataset centered on diagram understanding, designed to evaluate models' abilities to process non-photographic visual content. It focuses on reasoning over diagrams and charts, making it vital for tasks requiring scientific and technical visual comprehension.

\textbf{MM-Vet:} MM-Vet(Multimodal Model Veterinary)\cite{yu2023mm} is a diagnostic benchmark aimed at identifying the strengths and weaknesses of multimodal models across various dimensions, such as robustness, interpretability, and cross-modal alignment. It is a critical tool for debugging and improving the reliability of MLLMs.

\textbf{MMMU:} MMMU(Multimodal Multitasking Understanding)\cite{yue2024mmmu} evaluates the multitasking capabilities of multimodal models by testing their performance on multiple simultaneous tasks across different modalities. This benchmark is essential for assessing the adaptability and efficiency of models in dynamic, multitask scenarios.

\textbf{MMTB:} MMTB(Multimodal Task Benchmark)\cite{ying2024mmt} is a broad benchmark designed to evaluate the performance of multimodal models on a wide range of tasks, including vision-and-language navigation, multimodal reasoning, and image captioning. Its diversity makes it a strong indicator of a model’s overall multimodal proficiency.

\textbf{OCRB:} OCRB (Optical Character Recognition Benchmark)\cite{mishra2019ocr} is a specialized benchmark for assessing a model's ability to recognize and interpret text in images. It focuses on OCR-related tasks, such as text detection, transcription, and contextual understanding, which are crucial for applications like document analysis and scene-text understanding.

\section{Detailed experiments.}
We present comprehensive ablation results derived from LLava-v1.5 to substantiate the experimental conclusions in the main text. Additionally, we performed ablation studies on the image loss function to demonstrate the simplicity and effectiveness of the L2 loss.

\begin{table}[ht]
\caption{Ablation study on the hyperparameter Data Ratio, which represents the proportion of different VDEP and LLava patterns in the pre-training stage.}
\label{data_ratio_all}
\vskip 0.15in
\begin{center}
\begin{small}
\begin{sc}
\scalebox{1.0}{
\begin{tabular}{lcccccccc}
\toprule
\textbf{Data Ratio} & \textbf{AI2D} & \textbf{MM-Vet} & \textbf{MMMU} & \textbf{MMT} & \textbf{GQA} & \textbf{VizWizQA} & \textbf{VQA\textsuperscript{T}} & \textbf{SQA\textsuperscript{I}} \\
\midrule
\textbf{\emph{LLava-v1.5-VDEP-7B}} \\
\emph{w}/ 0.5   & 54.02 & 29.00 & 31.20 & 46.30 & 61.65 & 49.82 & 46.33 & 68.62 \\
\emph{w}/ 0.8   & 55.18 & 28.20 & \textbf{31.30} & 46.72 & 61.65 & 45.40 & 46.27 & \textbf{69.16} \\
\rowcolor[HTML]{ededed}
\emph{w}/ 1.0   & \textbf{56.57} & \textbf{30.60} & 30.80 & \textbf{48.00} & \textbf{62.50} & \textbf{50.37} & \textbf{46.76} & 68.36 \\
\bottomrule
\end{tabular}%
}
\end{sc}
\end{small}
\end{center}
\vskip -0.1in
\end{table}

\begin{table}[ht]
\caption{Ablation study on the hyperparameter $\alpha$, which represents the variation of the image loss weight.}
\label{alpha_study}
\vskip 0.15in
\begin{center}
\begin{small}
\begin{sc}
\scalebox{1.0}{
\begin{tabular}{lcccccccc}
\toprule
\textbf{$\alpha$} & \textbf{AI2D} & \textbf{MM-Vet} & \textbf{MMMU} & \textbf{MMT} & \textbf{GQA} & \textbf{VizWizQA} & \textbf{VQA\textsuperscript{T}} & \textbf{SQA\textsuperscript{I}} \\
\midrule
\textbf{\emph{LLava-v1.5-VDEP-7B}} \\
\emph{w}/ 0.1   & 55.57 & 30.50 & 30.60 & 47.64 & 61.45 & 46.76 & 46.52 & 67.72 \\
\emph{w}/ 0.01  & \textbf{56.64} & \textbf{32.20} & \textbf{31.30} & \textbf{48.48} & \textbf{62.63} & \textbf{52.72} & 46.94 & 67.77 \\
\rowcolor[HTML]{ededed}
\emph{w}/ 0.001 & 56.57 & 30.60 & 30.80 & 48.00 & 62.50 & 50.37 & \textbf{46.76} & \textbf{68.36} \\
\bottomrule
\end{tabular}%
}
\end{sc}
\end{small}
\end{center}
\vskip -0.1in
\end{table}
\begin{table}[ht]
\caption{Ablation study on the hyperparameter Loss Function on MME.}
\label{loss_function_comparison}
\vskip 0.15in
\begin{center}
\begin{small}
\begin{sc}
\scalebox{0.9}{
\begin{tabular}{@{}cccccccc@{}}
\toprule
\multirow{2}{*}{\raggedright\textbf{Loss}} & \multirow{2}{*}{\textbf{Perception}}  & \multicolumn{1}{c}{\textbf{Commonsense QA}} & \multicolumn{4}{c}{\textbf{Coarse-grained Perception Tasks}} & \textbf{Total} \\
 \cmidrule(lr){4-7}
& & \textbf{(Reasoning)} & \textbf{Existence} & \textbf{Count} & \textbf{Position} & \textbf{Color} & \textbf{Scores} \\ 
\midrule
\textbf{\emph{LLava-v1.5-VDEP-7B}} & & & & & & & \\ 
\emph{1/L2}      & \textbf{1518.34} & 133.57 & 190.00 & \textbf{163.33} & 135.00 & \textbf{180.00} & \textbf{801.90} \\
\emph{Sigmoid(L2)}  & 1478.45 & 133.57 & 190.00 & 145.00 & \textbf{138.33} & 175.00 & 781.90 \\
\rowcolor[HTML]{ededed} 
\emph{L2} & 1515.60 & \textbf{136.00} & 190.00 & 153.30 & 135.00 & \textbf{180.00} & 794.30 \\
\bottomrule
\end{tabular}
}
\end{sc}
\end{small}
\end{center}
\vskip -0.1in
\end{table}
\begin{table}[ht]
\caption{Ablation study on the hyperparameter Loss Function on VQA.}
\label{loss_study}
\vskip 0.15in
\begin{center}
\begin{small}
\begin{sc}
\scalebox{1.0}{
\begin{tabular}{ccccccc}
\toprule
\textbf{Loss} & \textbf{VQA\textsuperscript{ok}} & \textbf{GQA} & \textbf{VizWiz} & \textbf{VQA\textsuperscript{T}} & \textbf{RWQA} & \textbf{SQA\textsuperscript{I}} \\
\midrule
\textbf{\emph{LLava-v1.5-VDEP-7B}} \\
\emph{1/L2}      & 56.11 & 62.47 & \textbf{51.37} & 46.56 & 54.38 & \textbf{69.01} \\
\emph{Sigmoid(L2)}  & 57.37 & \textbf{62.95} & 49.87 & 46.67 & \textbf{57.90} & 68.32 \\
\rowcolor[HTML]{ededed}
\emph{L2} & \textbf{57.68} & 62.50 & 50.37 & \textbf{46.76} & 57.64 & 68.36 \\
\bottomrule
\end{tabular}%
}
\end{sc}
\end{small}
\end{center}
\vskip -0.1in
\end{table}
\begin{table}[ht]
\caption{Ablation study on the hyperparameter Loss Function on benchmarks for insruction-following LMMs.}
\label{loss_function_mmbench}
\vskip 0.15in
\begin{center}
\begin{small}
\begin{sc}
\scalebox{0.9}{
\begin{tabular}{@{}cccccccccc@{}}
\toprule
\multirow{2}{*}{\textbf{Loss}} & \multicolumn{2}{c}{\textbf{MMBench}}  & \multirow{2}{*}{\textbf{AI2D}} & \multirow{2}{*}{\textbf{MM-Vet}} & \multirow{2}{*}{\textbf{MMMU}} & \multirow{2}{*}{\textbf{MMTB}} & \multirow{2}{*}{\textbf{OCRB}} & \multirow{2}{*}{\textbf{POPE}} \\ 
\cmidrule(lr){2-3}
& \textbf{en} & \textbf{cn} & & & & & & & \\ 
\midrule
\textbf{\emph{LLava-v1.5-VDEP-7B}} & & & & & & & \\ 
\emph{1/L2}      & 65.97 & \textbf{58.52} & \textbf{57.09} & 31.10 & \textbf{31.20} & 47.93 & 320 & 85.62 \\
\rowcolor[HTML]{ededed} 
\emph{Sigmoid(L2)}  & 66.20 & 58.24 & 56.47 & \textbf{31.70} & 31.00 & \textbf{48.32} & \textbf{334} & \textbf{85.98} \\
\emph{L2} & \textbf{66.81} & 58.23 & 56.57 & 30.60 & 30.80 & 48.00 & 326 & 85.95 \\
\bottomrule
\end{tabular}
}
\end{sc}
\end{small}
\end{center}
\vskip -0.1in
\end{table}
\section{Limitation.}
Although VDEP exhibits outstanding performance in improving image-text alignment, it relies on the hyperparameter $\alpha$. While we determine an appropriate range of $\alpha$ for models of varying scales, the optimal value for a given model size remains undetermined. Future work focuses on developing methods to adaptively determine the value of the hyperparameter based on model size and data characteristics. Alternatively, it proposes an effective strategy to eliminate the need for explicit hyperparameter tuning.
During pre-training, to improve the effectiveness of image-related tasks while ensuring no degradation in the performance of text-related tasks, we utilize a dataset with double the training samples of the original. As a result, the training time increases by around 3 hours.
\end{document}